\documentclass[fleqn]{article} 
\usepackage{jfrStyle}
\usepackage{graphicx}
\graphicspath{{figures/}}
\usepackage{subfig}
\usepackage{apalike}
\let\bibhang\relax
\usepackage[utf8]{inputenc}
\usepackage{natbib}
\usepackage{setspace}
\usepackage{csvsimple}
\usepackage{lineno}
\usepackage{cite}
\usepackage{url}
\usepackage{hyperref}
\usepackage{siunitx}
\usepackage[margin=1in,footskip=.5in]{geometry}
\usepackage{setspace}
\usepackage{verbatim}
\usepackage{amssymb}
\usepackage{color}
\usepackage{amsmath}
\usepackage[toc]{appendix}
\usepackage{lscape}
\usepackage{authblk}
\usepackage{stmaryrd}
\usepackage{threeparttable}
\usepackage{booktabs}
\usepackage{multirow}
\usepackage[dvipsnames]{xcolor}
\usepackage{pifont}
\usepackage{xcolor}
\usepackage[most]{tcolorbox}
\newtcolorbox{highlighted}{colback=white,breakable}

\usepackage{lmodern}
\usepackage{mathtools, nccmath}

\newcommand{\rmse}[1]{{\tiny (#1)}}

\usepackage{xparse}
\DeclarePairedDelimiterX{\set}[1]{\{}{\}}{\setargs{#1}}
\NewDocumentCommand{\setargs}{>{\SplitArgument{1}{;}}m}
{\setargsaux#1}
\NewDocumentCommand{\setargsaux}{mm}
{\IfNoValueTF{#2}{#1} {#1\,\delimsize|\,\mathopen{}#2}}
\usepackage{nomencl}
\makenomenclature

\usepackage{etoolbox}
\renewcommand\nomgroup[1]{%
  \item[\bfseries
  \ifstrequal{#1}{S}{Symbols}{%
  \ifstrequal{#1}{A}{Abbreviations}{%
  \ifstrequal{#1}{C}{Other Symbols}{}}}%
]}

\usepackage{titlesec}
\titleformat{\paragraph}
{\normalfont\normalsize\bfseries}{\theparagraph}{1em}{}
\titlespacing*{\paragraph}
{0pt}{3.25ex plus 1ex minus .2ex}{1.5ex plus .2ex}
\usepackage[ruled,vlined]{algorithm2e}
\usepackage{array,tabularx,calc}

\title{\fontsize{16}{20}\selectfont PointRAFT: 3D deep learning for high-throughput prediction of potato tuber weight from partial point clouds}

\author[1]{Pieter M. Blok}
\author[1]{Haozhou Wang}
\author[2]{Hyun Kwon Suh}
\author[1]{Peicheng Wang}
\author[1]{James Burridge}
\author[1,*]{Wei Guo}

\affil[1]{Graduate School of Agricultural and Life Sciences, The University of Tokyo, Tokyo, Japan}
\affil[2]{Department of Integrative Biological Sciences and Industry, Sejong University, Seoul, Republic of Korea}
\affil[*]{Corresponding author. Email: guowei@g.ecc.u-tokyo.ac.jp}

\providecommand{\keywords}[1]
{
  \textbf{\textit{Keywords:}} #1
}

\begin{document}
\doublespacing

\maketitle

\begin{abstract}

Key phenotypic traits, such as crop size, volume, and weight, can be estimated non-destructively from point clouds. In practice, however, point clouds of crops are typically incomplete due to self-occlusion, leading to systematic errors when estimating these traits. To overcome this limitation, crop traits can be inferred directly from incomplete point clouds using deep learning-based regression networks. These networks, however, may perform suboptimally, because their predictions do not incorporate prior geometric knowledge of the crop. Such geometric knowledge is often available in structured settings, such as on a potato harvester, where the fixed distance between the camera and the conveyor belt allows tuber height to be easily derived from the point cloud. To exploit this geometric cue, we introduce PointRAFT, a high-throughput point cloud regression network that incorporates object height through a dedicated height embedding. We hypothesized that this height embedding improves tuber weight estimation from partial point clouds. To test this hypothesis, we collected 26,688 point clouds of 859 potato tubers on an operational harvester, spanning four cultivars and three growing seasons. On a test set of 5,254 point clouds representing 172 unique tubers, PointRAFT achieved a mean absolute error (MAE) of 11.7 g and a root mean squared error (RMSE) of 17.1 g, outperforming six point cloud regression networks without a height embedding. Requiring only 6.3 ms per point cloud for preprocessing and inference, PointRAFT’s regression stage is fast enough for high-throughput weight estimation on a commercial potato harvester. Beyond potato weight estimation, PointRAFT provides a versatile regression network applicable to a wide range of 3D phenotyping and robotic perception tasks. The code, network weights, and a subset of the dataset are publicly available at \url{https://github.com/pieterblok/pointraft.git}.
\end{abstract}


\clearpage

\section{Introduction}
\label{introduction}
Precision agriculture replaces uniform field management with site-specific interventions \citep{bullock2002, evert2017}. For potatoes (\textit{Solanum tuberosum}), these interventions rely on accurate yield data, which can be collected with in-line yield monitoring. In-line yield monitoring measures tuber weight directly on the harvester's conveyor belt, without diverting or handling the tubers. In current practice, this is done with load cells attached to the conveyor belt, which measure the mass of the harvested produce in real time \citep{zamani2014, kabir2018}. Load-cell systems are easy to use and maintain, but they measure gross mass. This gross mass includes tare, such as soil clods, stones, and plant residue, which leads to overestimation of tuber yield.

Camera-based yield monitoring systems are a more accurate alternative, because they can visually distinguish potato tubers from tare \citep{noordam2000, hofstee2003, elmasry2012, razmjooy2012, lee2018, long2018, si2018, su2018, pandey2019, cai2020, lee2020, dolata2021, huynh2022, jang2023, emwinghare2025}. Most systems rely on RGB color cameras and estimate tuber size, volume, or weight using pixel-to-world conversions that are calibrated under controlled laboratory conditions. Unfortunately, these externally calibrated conversions are prone to errors when the height profile of the tubers on the conveyor belt varies, which is the norm during practical harvesting.

RGB-Depth (RGB-D) cameras offer a promising solution. These cameras are inexpensive, easy to integrate, and generate both a color image and a depth image, which allows more accurate conversions. The color and depth images can also be fused into (colored) 3D point clouds from which tuber size, volume, or weight can be estimated using 3D algorithms. RGB-D-based point clouds, however, suffer from self-occlusion, meaning that the captured 3D shape is only partially observed due to the single viewpoint of the RGB-D camera \citep{fu2020}. Self-occlusion leads to underestimation of tuber size, volume, or weight, which is undesirable for yield monitoring. Multi-camera configurations can reduce this effect but cannot fully eliminate it, and their integration on a potato harvester remains challenging.

The partial observability of RGB-D point clouds has driven recent research interest in 3D shape completion algorithms \citep{fei2022}. Most shape completion algorithms use deep learning techniques to infer the full 3D geometry of an object from a partial observation. These methods are based on voxel-based networks, point cloud–based architectures, mesh-based models, or implicit representations \citep{fei2022}. Although many completion methods exist, including some developed for agricultural products \citep{ge2020, marangoz2020, chen2023, pan2023, xu2023, magistri2024, wangshape2025, zhang2025, wang2026}, only two methods currently offer the millisecond-level processing speed required for use on a potato harvester: CoRe \citep{magistri2022} and CoRe++ \citep{blok2025}.

CoRe and CoRe++ themselves face three challenges that limit their practical implementation on a potato harvester. First, training these methods requires large datasets of paired partial and complete point clouds, which are costly to obtain. The main bottleneck is the time-intensive generation of full 3D data using methods such as Gaussian Splatting or Structure-from-Motion, which typically takes about five minutes per potato. This is followed by time-consuming object registrations using approaches such as those described in \citet{wang2025}. Second, both CoRe and CoRe++ rely on 4-channel RGB-D images, which are sensitive to camera placement and tend to generalize less robustly than raw 3D point cloud data. Third, CoRe and CoRe++ require post-processing steps to convert their output point clouds into watertight meshes for volume extraction. This post-processing adds computational overhead and depends on mesh generation parameters that generalize poorly. 

Beyond 3D shape completion methods, another line of research has emerged that bypasses explicit 3D reconstruction entirely and performs direct regression on partial point clouds. With point cloud regression networks, a continuous target value, like volume or weight, can be directly estimated from 3D points using the network's dedicated regression head. Consequently, this eliminates the need for paired complete 3D shapes, improves generalizability across camera placements, removes the requirement for mesh-based post-processing, and directly estimates the weight, without relying on intermediate volume estimation. Point cloud regression networks have proven successful in estimating body weight from blanket-covered hospital patients \citep{bigalke2021} and human body fat percentage estimation \citep{hu2023}. In animal science, point cloud regression using PointNet \citep{qi2017pointnet} and PointNet++ \citep{qi2017pointnet++} has been successfully applied to estimate the weight of beef cattle \citep{hou2023}, pigs \citep{paudel2024}, and chickens \citep{zheng2025}. The same principle has been extended to forestry, where \citet{pourdelan2026} regressed plot-level wood volume and aboveground biomass directly from partially completed LiDAR point clouds. These studies demonstrate that complete geometry is not strictly necessary for accurate prediction of 3D shape properties. 

While the above literature demonstrates that point cloud regression can replace full 3D reconstruction, existing methods infer target values from the point cloud geometry alone. Under self-occlusion, they must therefore implicitly estimate the unobserved surface from the visible points, which is particularly challenging for small and irregular objects such as potato tubers. Yet, additional geometric knowledge is often available. On a potato harvester, for example, the fixed distance between the camera and the conveyor belt allows tuber height to be derived from the point cloud, which directly constrains the unobserved lower tuber surface.

To exploit this geometric cue, we introduce PointRAFT, a 3D regression network whose key architectural novelty is an object height embedding. This embedding feeds the approximate tuber height into the network, allowing PointRAFT to compensate for the unobserved lower tuber surface. We hypothesize that PointRAFT's height embedding improves weight estimation from partial point clouds, compared to regression networks that rely on the point cloud geometry alone. To test this hypothesis, we compare PointRAFT with six point cloud regression networks without a height embedding. This comparison is conducted on 26,688 point clouds of 859 potato tubers, acquired on an operational harvester across four cultivars and three growing seasons. Our main contribution is a point cloud regression network that incorporates prior geometric knowledge to better deal with self-occlusion. To support further research, we publicly released the PointRAFT code at \url{https://github.com/pieterblok/pointraft.git} and a subset of our dataset via \citet{wang2025}.
\clearpage

\section{Materials and methods}
\label{materials_methods} 

\subsection{PointRAFT}

\begin{figure}[hbt!]
  \centering
    \includegraphics[width=1\textwidth]{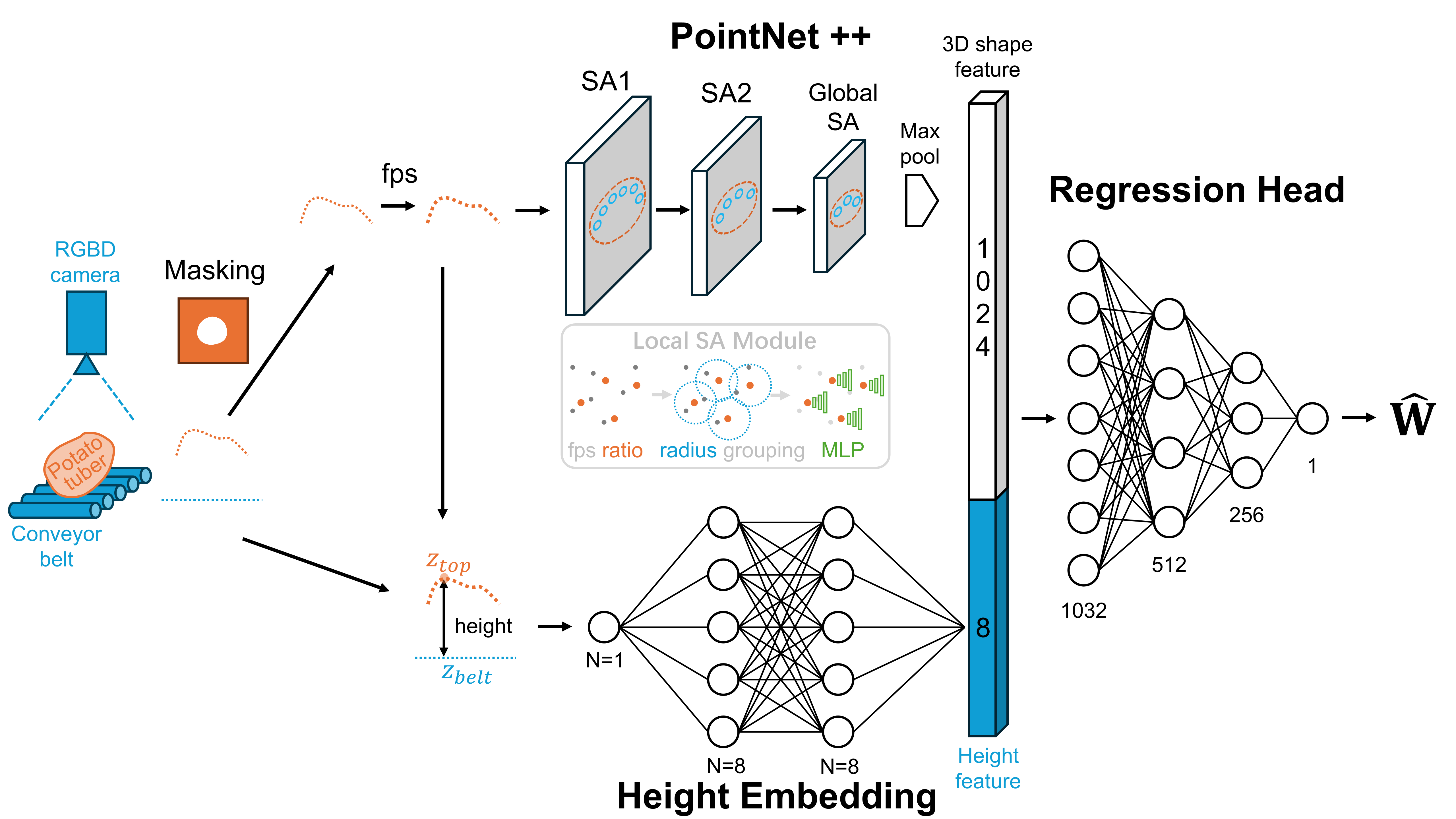}
    \caption{Schematic representation of the PointRAFT network, which directly regresses continuous 3D shape properties, such as tuber weight, from partial point clouds. Local geometric structures are captured by the set abstraction (SA) modules of PointNet++ (inset), which sample representative points, group their neighbors, and encode each neighborhood with a shared MLP. The architectural novelty of PointRAFT lies in the object height embedding, which provides an additional geometric cue for weight estimation.}
    \label{fig:pointraft_architecture}
\end{figure}

To estimate the weight of potato tubers from partial point clouds, we propose PointRAFT. While its overall architecture follows the hierarchical set abstraction (SA) principle of PointNet++ \citep{qi2017pointnet++}, we added a novel object height embedding to improve the weight regression. Additionally, a PyTorch-based point cloud downsampling method was implemented to improve sampling efficiency. The entire framework was developed using the PyTorch Geometric software library (version 2.7.0), and a schematic representation of the architecture is shown in Figure~\ref{fig:pointraft_architecture}.

\subsubsection{Preprocessing of partial point clouds}
\label{preprocessing}

Each potato tuber is represented by a 3D point cloud with a number of points that depends on the tuber's physical size and distance to the sensor. Larger tubers typically generate point clouds with several thousand points, while smaller tubers produce sparser clouds. To allow stable training and efficient batching, these raw point clouds were downsampled and preprocessed before being passed to the network.

A common downsampling method is farthest point sampling (FPS), which produces a spatially uniform subset of points and preserves the overall geometry of the object. FPS can be computationally expensive because it repeatedly computes distances across the full point cloud, making it a bottleneck for larger point clouds. To address this limitation, we implemented a PyTorch-based FPS downsampling method as a novel component of our pipeline, implemented using the pytorch\_fpsample software library \citep{han2023}. This software allowed efficient downsampling while retaining the geometric advantages of FPS. In our implementation, each tuber was downsampled to a fixed set $\mathbf{P}_0$ of 1024 points (Equation~\ref{eq_pcd}).

\begin{equation}
\label{eq_pcd}
\mathbf{P}_0 = \{ (x_i, y_i, z_i) \} \in \mathbb{R}^{N \times 3} \quad \text{for} \quad i = 1, 2, \dots, N
\end{equation}

In addition to downsampling, we applied PyTorch Geometric's \texttt{Center} transform to each point cloud, shifting all points so that the centroid of the transformed cloud aligns with the origin (0,0,0) of the camera coordinate frame. This ensured that each point cloud was centered consistently, eliminating potential biases caused by the absolute positions of tubers in the sensor’s coordinate frame. We did not normalize the scale of the point clouds to preserve the relative sizes of the tubers. This ensured that larger tubers remained larger in the network’s representation, allowing the model to learn geometric cues related to size. The centering and PyTorch-based FPS downsampling formed the point cloud preprocessing module of PointRAFT.

\subsubsection{Neural network architecture}

The backbone of PointRAFT was based on the PointNet++ single scale grouping (SSG) architecture \citep{qi2017pointnet++}. PointNet++ SSG was chosen for its efficiency and its ability to encode hierarchical geometric features using a series of hierarchical set abstraction (SA) modules. Each local SA module follows three main steps: selecting representative points by FPS, grouping nearby points within a specified radius, and learning geometric features with a shared MLP. This structure is illustrated in the inset of Figure~\ref{fig:pointraft_architecture}.

The first SA module downsampled the input point cloud $\mathbf{P}_0$ by selecting 50\% of its points using farthest point sampling. The resulting subset is denoted as $\mathbf{P}_1$. For each point in $\mathbf{P}_1$, up to 64 neighboring points within a radius of 0.2 m were grouped for feature learning. The radius was chosen to be larger than the physical dimensions of individual potato tubers, so that each local region captured the entire partial point cloud. This was chosen because it enabled object-level feature learning in a hierarchical structure, which was deemed beneficial for self-occluded partial point clouds. Geometric features were then learned using a PointNet-style convolution implemented with shared multi-layer perceptrons (MLPs) with layer sizes [3, 64, 64, 128], producing a 128-dimensional feature vector for each sampled region in $\mathbf{P}_1$.

The second SA module further reduced the point set 
$\mathbf{P}_1$ to 25\% of its points using farthest point sampling, resulting in the point subset $\mathbf{P}_2$. For each point in $\mathbf{P}_2$, up to 64 neighboring points within a radius of 0.4 m were selected from $\mathbf{P}_1$. This radius, again larger than the physical dimensions of individual tubers, allowed the network to learn object-level features from a coarser representation of the points. The input to the MLP at this stage had dimensionality 128+3, where the three additional channels corresponded to the relative 3D coordinates of each selected neighboring point in $\mathbf{P}_1$, concatenated with the learned features from the previous SA layer. Shared MLPs with layer sizes [131, 128, 128, 256] then produced a 256-dimensional feature vector for each sampled region in $\mathbf{P}_2$, encoding coarser geometric features of the overall shape of the tuber.

The final global SA module used a simpler structure without FPS or radius-based grouping. It concatenated the 256-dimensional features from the second local SA module with the corresponding 3D coordinates in $\mathbf{P}_2$, yielding 259-dimensional point features. A shared MLP with layer sizes [259, 256, 512, 1024] transformed each point feature, after which global max pooling produced a single 1024-dimensional permutation-invariant global 3D shape feature vector.

To enhance the global 3D shape representation, we implemented a novel object height embedding that encoded the approximate height of each potato tuber. Tuber height can serve as an additional proxy for the tuber’s weight, which can improve neural network regression. Unfortunately, tuber height cannot be reliably observed from self-occluded partial point clouds, as the bottom parts of the tuber cannot be observed. Therefore, we embedded the tuber height by approximating it as the difference between the 3D point of the tuber closest to the camera and the known distance to the conveyor belt, assuming that most tubers lie flat on the conveyor belt. The approximated tuber height was encoded using a single MLP with layer sizes [1, 8, 8], producing an 8-dimensional feature vector. This vector was then concatenated with the 1024-dimensional global feature vector learned by the SA modules, resulting in a combined 1024+8-dimensional vector that captured both 3D shape and tuber height information. The combined vector was then processed by the regression head, implemented as a single MLP with layer sizes [1032, 512, 256, 1] and dropout of 0.5, to predict the tuber weight as a continuous value. 

\subsection{Dataset}
\label{data_collection}

\subsubsection{Imaging systems}
\label{imaging_system}
\begin{figure}[hbt!]
  \centering
  \subfloat[] {\includegraphics[width=1\textwidth]{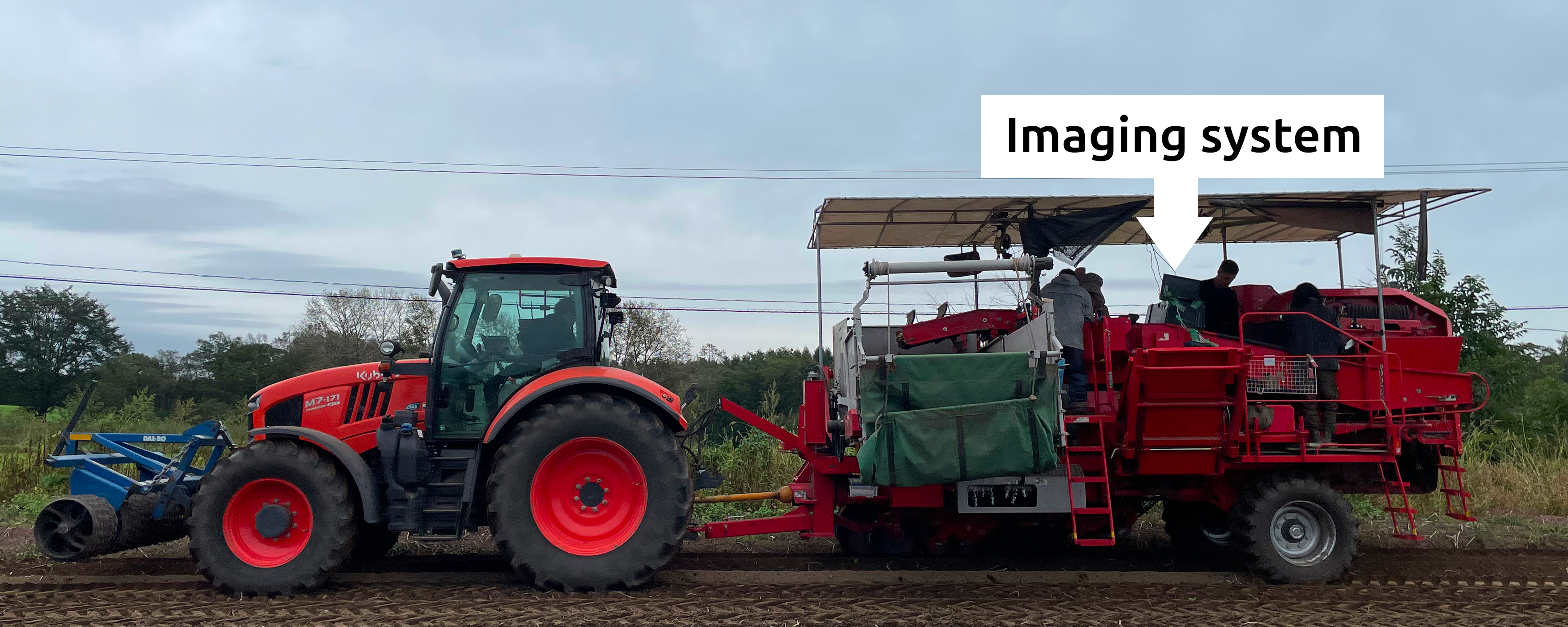}\label{fig:harvester}}
  \hfill
  \subfloat[] {\includegraphics[width=0.53\textwidth]{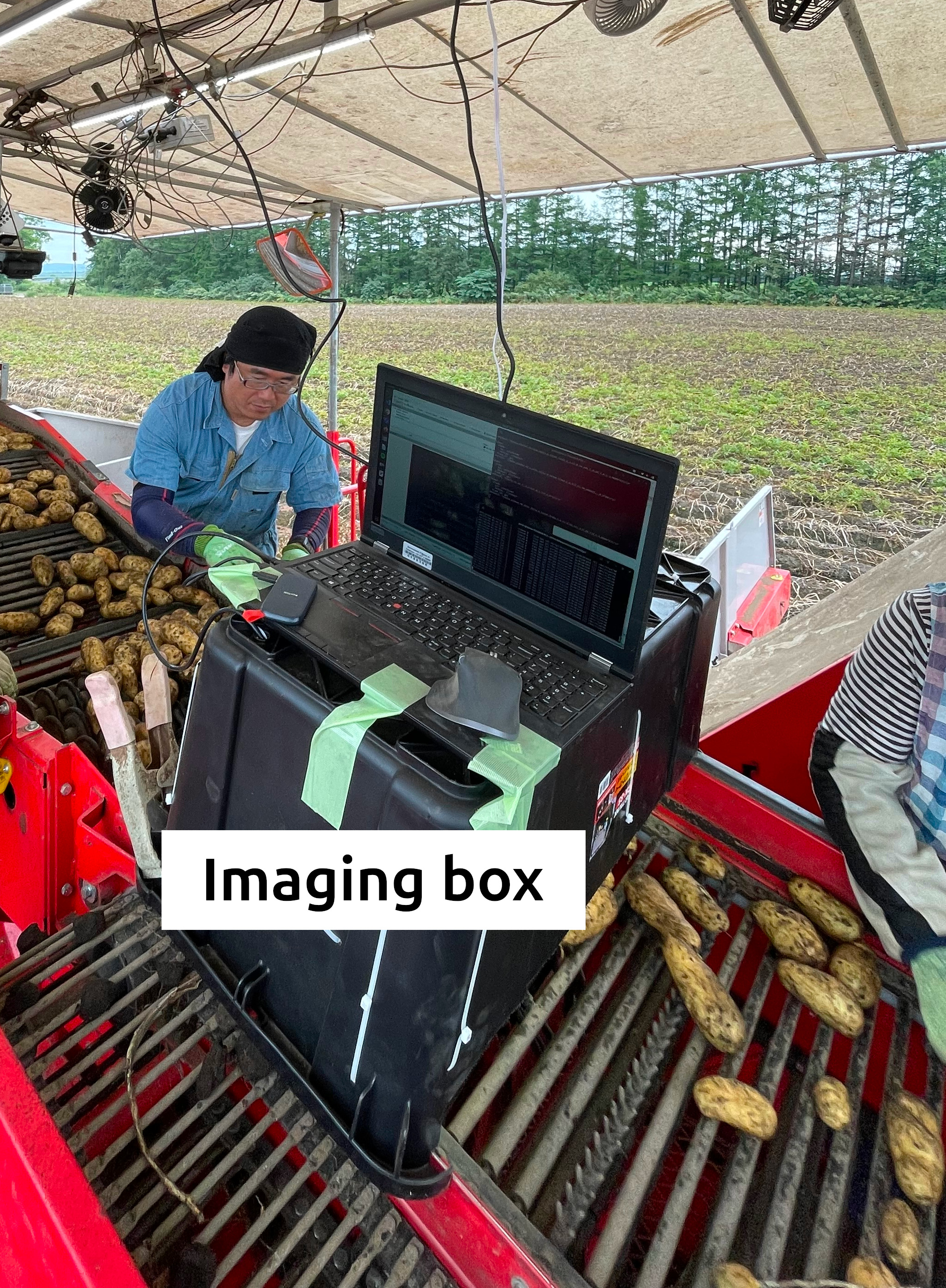}\label{fig:box_on_harvester}}
  \hfill
  \subfloat[] {\includegraphics[width=0.397\textwidth]{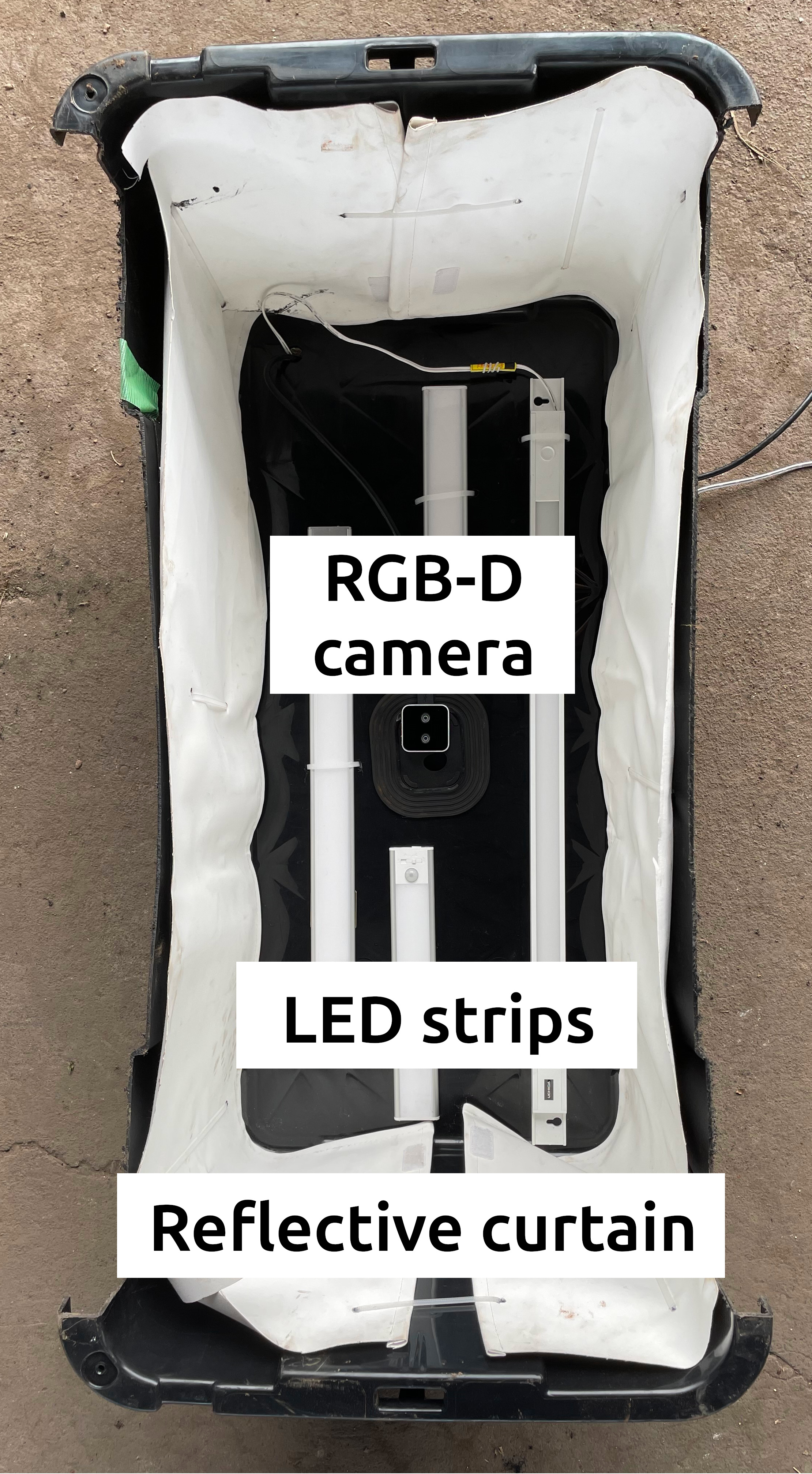}\label{fig:box_inside}}
  \caption{(a) and (b) provide overviews of the first imaging system, which was installed on a single-row potato harvester and used during the 2023 growing season. (c) Inside the imaging box, an RGB-D camera was installed, together with four LED strips that provided the necessary illumination inside the box. The sides of the box were covered with a reflective curtain to diffuse the light evenly across the conveyor belt.}
  \label{fig:data_collection_system}
\end{figure}

Our point cloud dataset was acquired using two RGB-D imaging systems, both installed above the conveyor belt of a single-row potato harvester (Toyonoki Top-1, Figure~\ref{fig:harvester}). The two systems shared the same RGB-D camera and data acquisition software, but differed in enclosure design, camera-to-belt distance, and field of view.

The first imaging system was constructed as a closed black plastic enclosure mounted above the conveyor belt, with dimensions 0.85 m (length), 0.45 m (depth), and 0.39 m (height), refer to Figure~\ref{fig:box_on_harvester}. Inside the enclosure, four LED strips with a color temperature of 6000 K were installed. Reflective curtains along the inner sides helped to diffuse the light evenly across the conveyor belt (Figure~\ref{fig:box_inside}). An RGB-D camera (Intel RealSense D405) was mounted centrally inside the enclosure, capturing a top-view perspective of the conveyor belt. The distance between the camera and the conveyor belt was approximately 0.35 m, resulting in a field of view of about 0.39 m (length) by 0.64 m (width). The camera's exposure time was set to 5.0 ms to avoid motion blur caused by the fast-moving potato tubers.

\begin{figure}[hbt!]
  \centering
  \subfloat[]{%
    \includegraphics[width=0.635\textwidth]{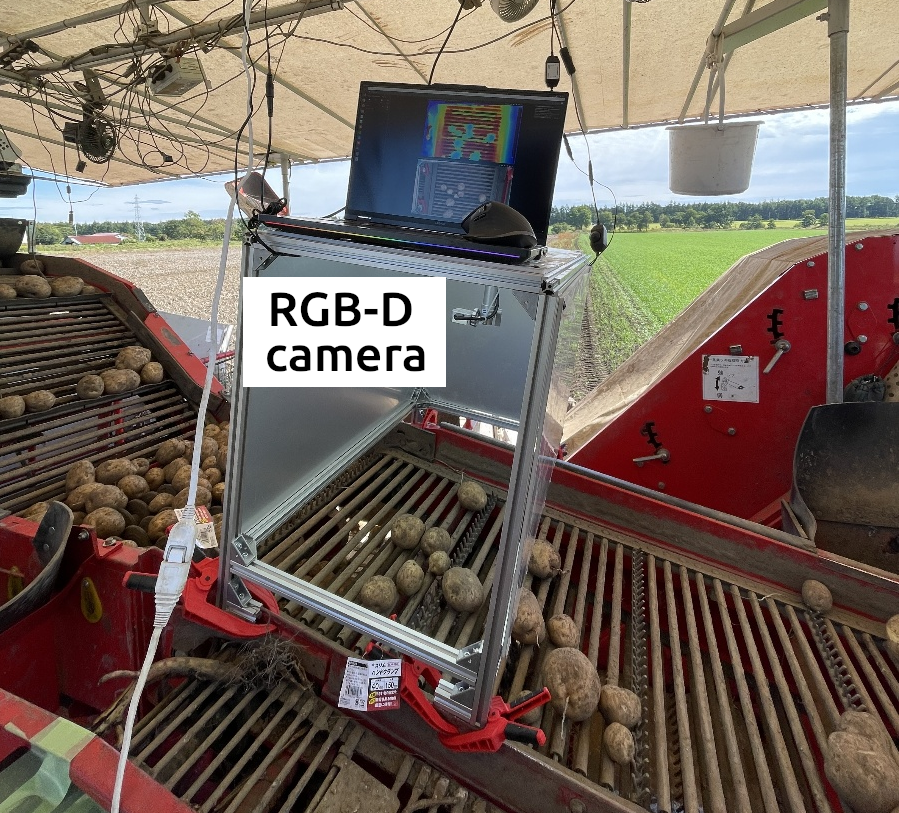}%
    \label{fig:system_01}%
  }\hfill
  \begin{minipage}[b]{0.325\textwidth}
    \centering
    \subfloat[]{%
      \includegraphics[width=\linewidth]{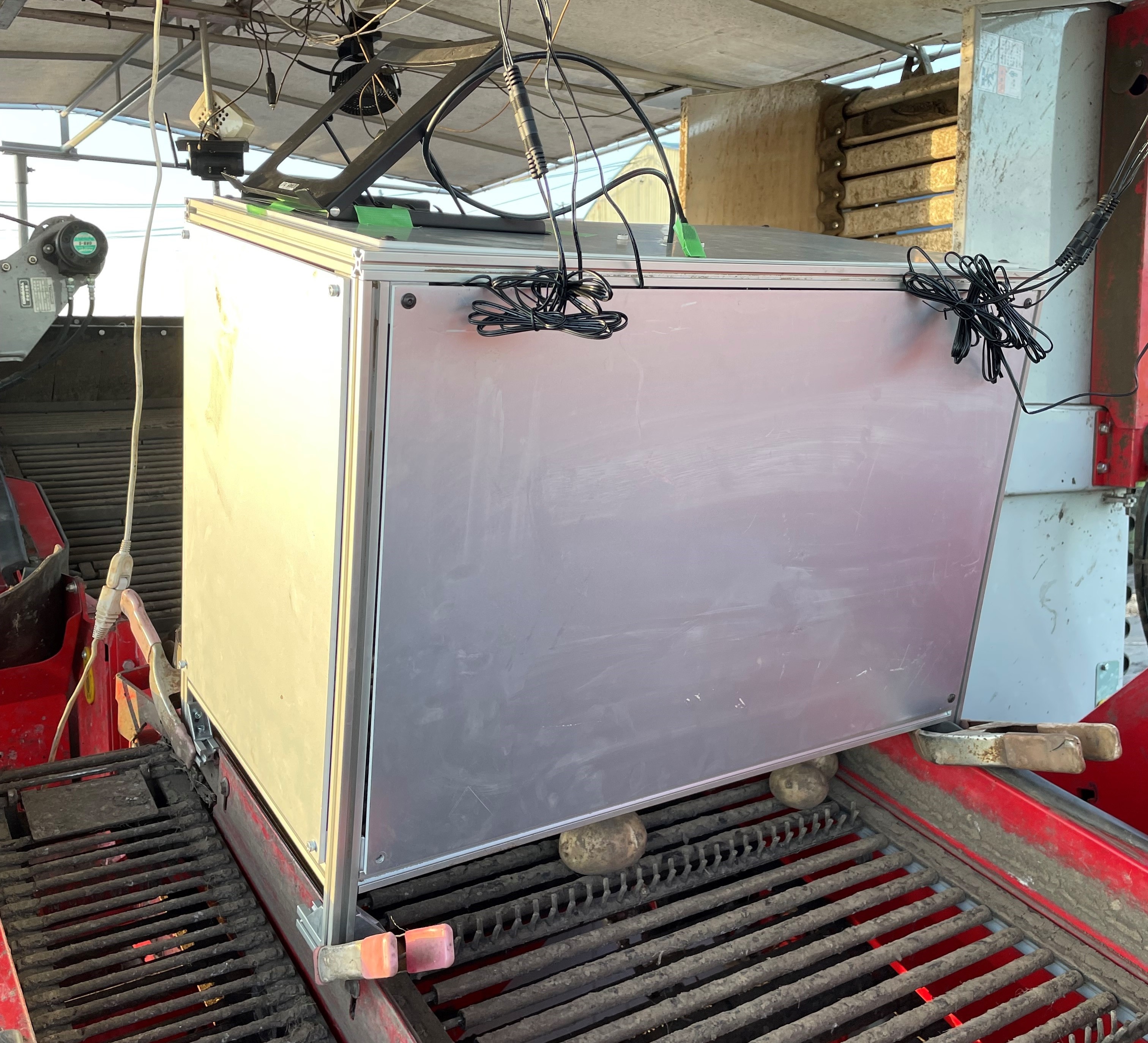}%
      \label{fig:system_02}%
    }\\[1ex]
    \subfloat[]{%
      \includegraphics[width=\linewidth]{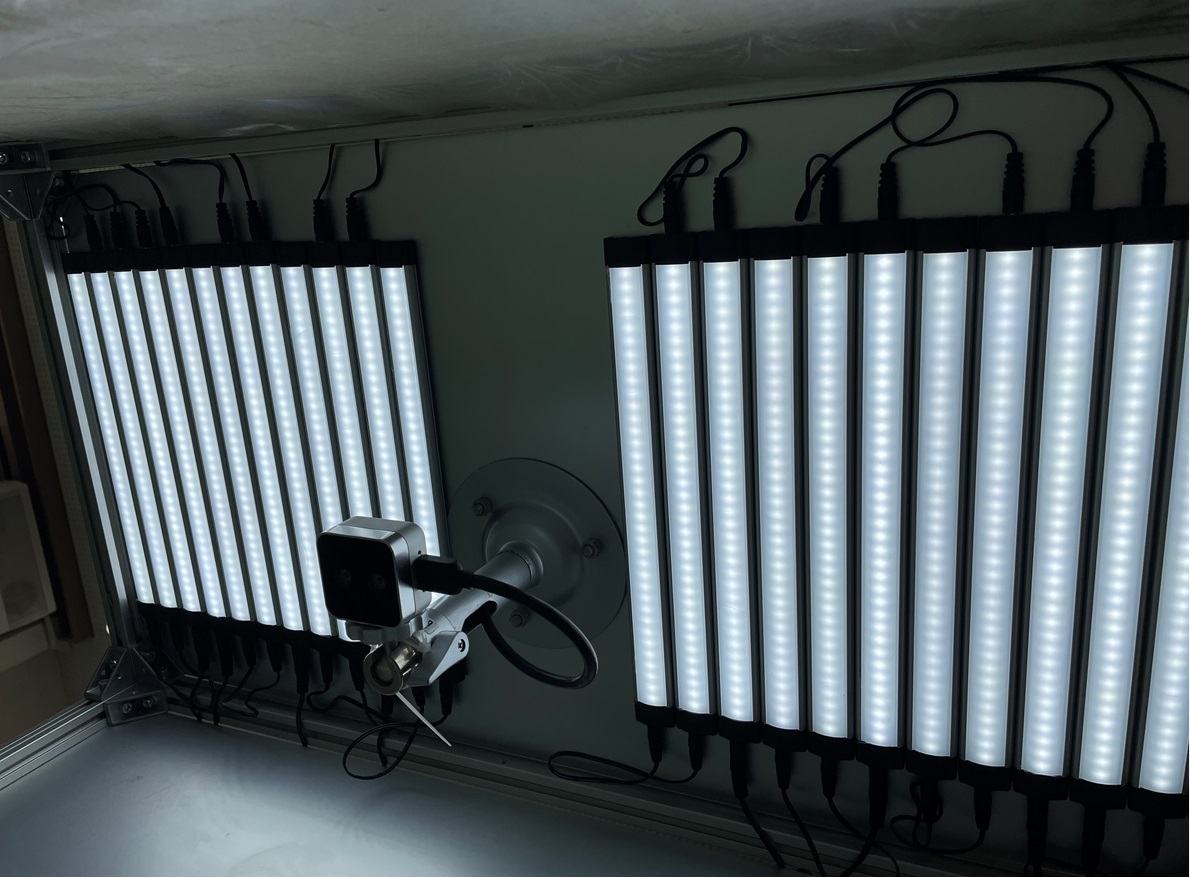}%
      \label{fig:system_03}%
    }
  \end{minipage}
  \caption{(a) Overview of the second imaging system, which was used during the 2024 and 2025 growing seasons. One side plate has been removed to reveal the internal components. (b) The yield monitoring system with enclosed sides during operation on the harvester. (c) On the ceiling plate of the enclosure, 24 LED strips and the RGB-D camera were installed.}
  \label{fig:system}
\end{figure}

The second imaging system was constructed using aluminum T-shaped profiles, forming a rigid enclosure with dimensions 0.85 m (length), 0.48 m (width), and 0.58 m (height) (Figure~\ref{fig:system_01}). The sides of the enclosure were covered with plates to reduce the influence of natural sunlight on the captured images (Figure~\ref{fig:system_02}). A total of 24 LED strips were mounted on the ceiling plate, providing uniform artificial illumination for measurements under both day and night conditions (Figure~\ref{fig:system_03}). Each LED strip was 30 cm long (AIBOO LED) with a color temperature of 6000 K. The same Intel Realsense D405 RGB-D camera was mounted centrally between the LED strips and manually aligned with the conveyor belt. The camera-to-belt distance was approximately 0.46 m, resulting in a field of view of about 0.52 m (length) by 0.75 m (width). As with the first system, the camera's exposure time was fixed at 5.0 ms to ensure motion-blur-free image acquisition. 


In both imaging systems, the camera streamed color and depth images at 30 frames per second via Robot Operating System 2 (ROS2, version: Humble Hawksbill) connected to a laptop computer (Lenovo Legion Pro 7 16IRX8H equipped with an Intel i9-13900HX CPU, 32 GB of RAM, and an NVIDIA GeForce RTX 4090 Laptop GPU with 16 GB of VRAM). The color and depth images were stored in ROS2 bag files. 

\subsubsection{Data collection on the harvester}
RGB-D image acquisition was carried out over three growing seasons: 2023, 2024, and 2025. During the 2023 season, data were collected with the first imaging system (Figure~\ref{fig:box_on_harvester}), while the second imaging system (Figure~\ref{fig:system_01}) was used for data acquisition in the 2024 and 2025 seasons. 

Image data were recorded in three potato fields located in Sarabetsu, Japan. The potato cultivars in the fields were Sayaka and Kitahime. During the 2023 season, images from two other cultivars, Corolle and Haruka, were collected by operating the harvester indoors, where boxes of potatoes were manually dumped onto the conveyor belt. As a result, the images collected in this study exhibited a wide range of tuber shapes and sizes, with each cultivar showing distinct visual characteristics. Haruka tubers were the most elongated, Corolle were moderately elongated, and Sayaka and Kitahime were comparatively rounder. In 2025, a fraction of Kitahime tubers were deformed due to the hot and dry growing conditions in Sarabetsu, which negatively affected tuber development and shape. Overall, the diversity of tuber shapes across the four cultivars, combined with the deformities observed in Kitahime, provided a rich and diverse dataset for analyzing the performance of the PointRAFT weight regression. 

During both field harvesting and indoor operation, 859 potato tubers of various sizes and shapes were collected from the conveyor belt of the potato harvester. The tubers were collected according to the following procedure: a person positioned in front of the imaging system randomly selected a potato tuber from the conveyor belt and inserted a colored thumbtack into it. The tack was inserted so that it remained visible in the RGB images as the tuber moved through the camera’s field of view. Based on the conveyor belt speed, each tagged potato appeared in approximately 20–35 consecutive image frames. After image recording in a ROS2 bag file, a second person positioned behind the imaging system collected the tagged tuber and placed it into a uniquely labeled paper bag. All samples were transported to the barn, where the tubers were weighed using a digital scale. Detailed information on the acquired datasets is presented in Table~\ref{tab:fielddata}.

\begin{table}[hbt!]
\captionsetup{justification=raggedright,singlelinecheck=false} 
    \caption{Overview of the three potato fields and the indoor experiment from which the data was acquired.}
    \centering
    \begin{tabular}{c c c c c c c c c}
        \hline
        & & & & Camera-belt & Potato & Sampled &  \\ 
        Year & Field & Latitude & Longitude & distance [m] & cultivar & potatoes & Note \\ \hline
        2023 & 1 & 42.610316 & 143.156753 & 0.35 & Sayaka & 280 &\\
        2023 & - & - & - & 0.35 & Kitahime & 31 & Indoor\\
        2023 & - & - & - & 0.35 & Corolle & 30 & Indoor\\
        2023 & - & - & - & 0.35 & Haruka & 30 & Indoor\\
        2024 & 2 & 42.616872 & 143.174312 & 0.46 & Kitahime & 258 &\\
        2025 & 3 & 42.600608 & 143.174741 & 0.46 & Kitahime & 230 &\\
        \hline
        \multicolumn{6}{l}{\textbf{Total}} & \textbf{859} &\\
        \hline
    \end{tabular}
    \label{tab:fielddata}
\end{table}

\subsubsection{Point cloud creation}
\label{point_cloud_creation}
After image acquisition, the ROS2 bag files were processed to extract individual RGB and depth frames corresponding to the sampled potato tubers. The selected RGB frames were annotated using LabelMe software (version 5.9.1), with assistance from the Segment Anything 2 model (SAM2) to increase annotation accuracy and speed. Once the tubers were annotated, the corresponding masks were clipped onto the RGB and depth frames. 

To generate the point clouds from the clipped RGB and depth frames, the Open3D software library (version 0.19.0) was used, along with the intrinsic camera parameters of the Intel RealSense D405. The obtained point clouds were post-processed to remove distant, low-density clusters of points that contained fewer points and were spatially separated from the main cluster representing the potato tuber. The filtering was performed using a custom method that applied the DBSCAN clustering algorithm to identify and eliminate these irrelevant clusters. In total, 26,688 point clouds were extracted from the 859 potato tubers, averaging 31 point clouds per tuber.

\subsubsection{Dataset splits}
\label{dataset_splits}
\begin{figure}[hbt!]
  \centering
    \includegraphics[width=1\textwidth]{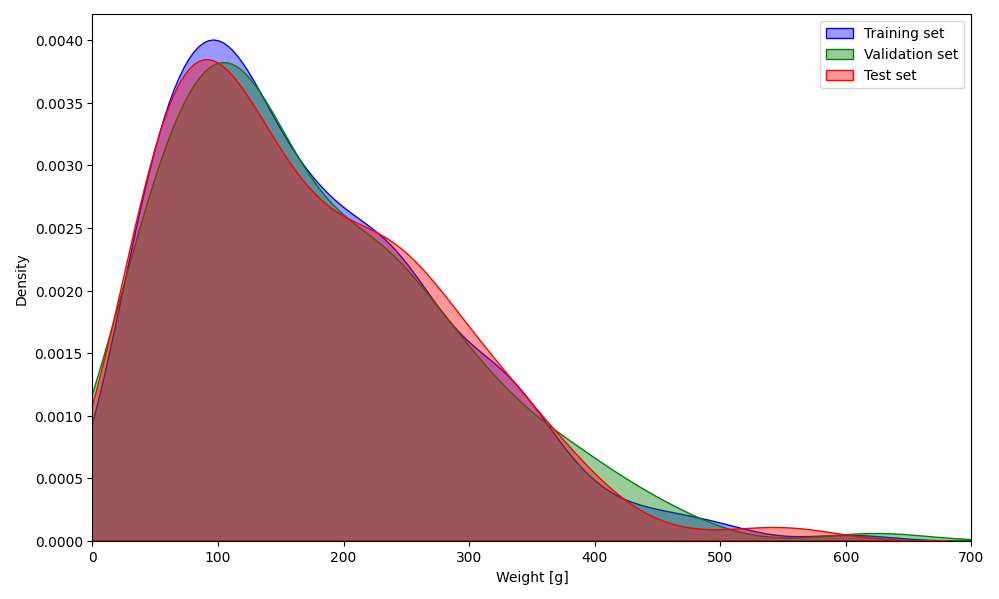}
    \caption{Kernel density estimate plot for visualizing the weight distribution in the training, validation, and test set.}
    \label{fig:histogram}
\end{figure}

The acquired dataset was split into three subsets for training, validation, and independent testing. To ensure a balanced representation of potato weights across the dataset splits, we applied stratified sampling. First, the weights of the 859 sampled potatoes were grouped into ten bins, each containing a similar number of samples (ranging from 83 to 88 potatoes per bin). The bin edges were set at 16~g, 51~g, 74~g, 98~g, 121~g, 150~g, 185~g, 224~g, 262~g, 325~g and 625~g. The dataset was then split with stratified labels into a training set of 515 tubers (60\%), a validation set of 172 tubers (20\%), and a test set of 172 tubers (20\%). The total number of point clouds belonging to the training samples was 16,108, the total for validation samples was 5,326, and the total for testing samples was 5,254. Despite using stratified sampling, the dataset remained heavily imbalanced, with smaller and larger tubers underrepresented relative to moderately sized tubers (50--250~g), refer to Figure~\ref{fig:histogram}.

\subsection{Neural network training}
\subsubsection{Embedding of tuber height}
\label{data_heightembedding}
Prior to applying the PyTorch Geometric \texttt{Center} transform to our point clouds (see Section~\ref{preprocessing}), the height of each tuber was estimated by subtracting the z-coordinate of the 3D point in $\mathbf{P}_0$ closest to the camera from the fixed camera-to-belt distance (Table~\ref{tab:fielddata}). The approximate height was included as an additional feature in PyTorch Geometric’s in-memory dataset class (\texttt{InMemoryDataset}). This class allowed efficient storage of all point clouds and their associated features in memory, enabling fast access during training and minimizing disk input/output overhead.

\subsubsection{Data augmentation}
\label{data_augmentation}
The \texttt{InMemoryDataset} also incorporated the data augmentation pipeline, enabling all transformations to be applied on-the-fly during batch loading. The augmentation pipeline consisted of seven transformations, which were applied during training to enhance the generalization performance of PointRAFT.

The first transformation was a random jitter of 5$\cdot$10\textsuperscript{-4}, which added small noise to the points’ 3D positions to simulate sensor inaccuracies and noise. Three random rotations were applied along the x, y, and z axes, with each axis rotated by random angles between 0° and 2°, simulating different viewing perspectives. Two random flips were applied along the x and y axes with a 50\% probability, further diversifying the possible orientations. Finally, a random shear transformation with a shear factor of 0.2 was applied across all three axes to simulate geometric distortions.

\subsubsection{Training details}
\label{training_details}
Instead of training PointRAFT with a fixed set of hyperparameters, the Optuna software framework (version 4.6.0) was employed to automatically identify an effective hyperparameter set based on validation performance. The optimization process explored batch sizes ranging from 16 to 64 in increments of 16, learning rates selected from {1$\cdot$10\textsuperscript{-2}, 5$\cdot$10\textsuperscript{-3}, 1$\cdot$10\textsuperscript{-3}, 5$\cdot$10\textsuperscript{-4}, 1$\cdot$10\textsuperscript{-4}}, and weight decay values from {1$\cdot$10\textsuperscript{-2}, 1$\cdot$10\textsuperscript{-3}, 1$\cdot$10\textsuperscript{-4}, 1$\cdot$10\textsuperscript{-5}}.

In addition, Optuna evaluated multiple loss functions, including mean squared error (MSE), mean absolute error (L1), and Smooth L1 loss. All Optuna trials were trained using the Adaptive Moment Estimation (Adam) optimizer for 50 epochs, with an exponentially decaying learning rate scheduler applied using a decay factor of 0.97.

The best-performing hyperparameter set consisted of a learning rate of 5$\cdot$10\textsuperscript{-4}, a weight decay of 1$\cdot$10\textsuperscript{-4}, a batch size of 32, and the Smooth L1 loss function (Equation~\ref{eq_smoothl1}). The $\beta$ value of the Smooth L1 loss function was set to 20.0, meaning that errors smaller than 20 were treated quadratically, like MSE, to emphasize accurate predictions for most samples, while errors larger than 20 were treated linearly, like L1, preventing extreme outliers from dominating training. Using these hyperparameters, PointRAFT was trained.

\begin{equation}
\label{eq_smoothl1}
\mathcal{L}_{\text{SmoothL1}}(W_i, \hat{W}_i) =
\begin{cases}
\frac{0.5\cdot(W_i - \hat{W}_i)^2}{\beta}, & \text{if } |W_i - \hat{W}_i| < \beta, \\
|W_i - \hat{W}_i| - 0.5 \cdot \beta, & \text{otherwise},
\end{cases}
\end{equation}

where $W_i$ denotes the ground truth weight, and $\hat{W}_i$ the predicted weight. 

\subsection{Evaluation}
\label{evaluation}
\subsubsection{Benchmark analysis}
\label{benchmark_analysis}
To assess the performance of PointRAFT in estimating tuber weight from partial point clouds, we compared it against six benchmark regression networks: PointNet \citep{qi2017pointnet}, PointNet++ \citep{qi2017pointnet++}, DGCNN \citep{wang2019dgcnn}, PointMLP \citep{ma2022}, RandLA-Net \citep{hu2020randlanet}, and Point Transformer V3 (PTv3) \citep{wu2024ptv3}. Since these networks were originally designed for classification or segmentation, their output layers were replaced by a global pooling operation and a regression head that predicts one weight per point cloud. PointNet++ shares its backbone with PointRAFT, but lacks the object height embedding. For PTv3, which was developed for large-scale scenes, we used a compact version with three encoder stages to suit the size of a single tuber, referred to as PTv3-C. Implementation details of all benchmarks are available in our code repository. All neural networks used the same tuber-level data split, 1,024 centered XYZ input points, 50-epoch training budget, optimizer, scheduler, loss function, and evaluation procedure. None of the comparison networks had the object height embedding.

PointRAFT and the six benchmark models were assessed using the Mean Absolute Error (MAE, Equation~\ref{eq_mae}), the Root Mean Squared Error (RMSE, Equation~\ref{eq_rmse}), the coefficient of determination ($R^2$, Equation~\ref{eq_r2}), and the bias (Equation~\ref{eq_bias}). The MAE gives the average deviation per tuber and therefore directly indicates the error that propagates into the estimated yield. The RMSE penalizes large deviations, so a much higher RMSE than MAE indicates that a model is accurate for most tubers but fails on a few. The $R^2$ expresses how much of the variation in tuber weight is explained by the predictions, where a value of 1 indicates a perfect fit. The bias is the mean signed error and shows whether a model systematically overestimates (positive bias) or underestimates (negative bias) the tuber weight. Because positive and negative errors cancel each other out, a bias close to zero does not imply accurate individual predictions, but indicates that the errors are balanced.

\begin{equation}
\label{eq_mae}
\text{MAE} = \frac{1}{N} \sum_{i=1}^{N} \left| W_i - \hat{W}_i \right|
\end{equation}

\begin{equation}
\label{eq_rmse}
\text{RMSE} = \sqrt{ \frac{1}{N} \sum_{i=1}^{N} \left( W_i - \hat{W}_i \right)^2 }
\end{equation}

\begin{equation}
\label{eq_r2}
R^2 = 1 - \frac{\sum_{i=1}^{N} \left( W_i - \hat{W}_i \right)^2}{\sum_{i=1}^{N} \left( W_i - \overline{W} \right)^2}
\end{equation}

\begin{equation}
\label{eq_bias}
\text{Bias} = \frac{1}{N} \sum_{i=1}^{N} \left( \hat{W}_i - W_i \right)
\end{equation}

where $N$ denotes the total number of test samples, $W_i$ the corresponding ground truth weight, $\hat{W}_i$ the predicted weight, and $\overline{W}$ the mean ground truth weight of the test samples.

The above metrics were calculated per point cloud. Since each test tuber was represented by approximately 31 point clouds, these point clouds were not independent from each other. Moreover, a yield monitoring system requires only one weight estimate per tuber. For this reason, the MAE and RMSE of the entire test set were also calculated per tuber, by first averaging the predictions of all point clouds of the same tuber. To test whether PointRAFT performed significantly better than the other networks, a paired Wilcoxon signed-rank test was applied to the absolute errors of the 172 test tubers, with a significance level of 0.05.

Besides the performance on the entire test set, the weight estimation was analyzed for three types of subsets to assess how consistently the networks performed under practical harvesting conditions. The first subset analysis concerned the potato cultivar, since a weight estimation method should perform consistently across cultivars. The second subset analysis concerned the growing season, indicating whether the networks remained accurate for tubers from the different growing conditions and camera setups in our dataset. The third subset analysis concerned the tuber weight, since tubers of widely varying weights pass over the harvester's conveyor belt. For this purpose, the test tubers were divided into six weight classes of 100~g based on their ground truth weight, ranging from 0--100~g to 500--600~g, where each class included its upper bound.

Next, we analyzed in which horizontal image region the weight estimation should ideally be performed. The reason for this analysis is that it is neither computationally desirable nor necessary to analyze all approximately 31 RGB-D frames of each tuber when it moves over the conveyor belt. The computational efficiency can be improved by performing the weight regression only in the image region with the average lowest MAE. We therefore examined eleven horizontal image regions, defined by the vertical pixel position of the tuber in the 720-pixel-high image, counted from the bottom of the image, where the tubers enter the field of view. The first region covered 0--150 pixels, followed by nine regions of 50 pixels each between 150 and 600 pixels, and a last region covering 600--720 pixels. The two outer regions were larger because tubers were rarely captured near the bottom and top edges of the image. For each region, the MAE of PointRAFT and the other benchmarks was calculated over all test point clouds within that region.

Analyzing only one point cloud per tuber reduces the computational load, but the network must still process up to 50 tubers that can be present on the conveyor belt at the same time, which requires a short processing time per point cloud. To evaluate whether PointRAFT and the other benchmarks meet this high-throughput requirement, we calculated the average analysis time per point cloud on the point clouds of the test set. This analysis was conducted using the same laptop as described in Section~\ref{imaging_system}. Since our original \texttt{InMemoryDataset} class already stored the preprocessed point clouds, we implemented a custom timing script that performed the complete per-cloud processing: point cloud loading, height estimation, centering, farthest point sampling to the required number of input points, and the forward pass with a batch size of one. To obtain stable measurements, each network was first warmed up on 50 point clouds, after which all point clouds of the test set were processed three times and the mean processing time per point cloud was reported. The reported processing times excluded RGB-D point cloud creation and DBSCAN filtering (Section~\ref{point_cloud_creation}).

\subsubsection{Ablation study}

An ablation study was conducted to quantify the contribution of the individual components of PointRAFT to understand their influence on the weight estimation performance. Eleven ablations were evaluated relative to the PointRAFT configuration described earlier, using identical training and evaluation protocols, including the inference time measurement described in Section~\ref{benchmark_analysis}.

The first two ablations focused on the input point cloud resolution, as this directly affects geometric detail and computational cost. To study the trade-off between performance and throughput, the network was trained and evaluated with 512 and 2048 input points, instead of the default 1024. This analysis allowed us to evaluate whether fewer points are sufficient for reliable regression and whether denser point clouds provide additional benefits for self-occluded tubers.

The next three ablations examined training-related choices. In the first of these, PointRAFT was trained without data augmentation to study its influence on weight regression. The loss function was then varied by replacing the Smooth L1 loss with L1 and MSE loss functions, allowing us to assess the sensitivity of the regression task to less and more outlier-sensitive error formulations.

Finally, a series of six ablations examined the impact of PointRAFT's main architectural novelty, the object height embedding. The width of the MLP used for height encoding was varied by testing two narrower configurations, [1,2,2] and [1,4,4], and four wider configurations, [1,16,16], [1,32,32], [1,48,48] and [1,64,64], relative to the default height embedding of [1,8,8]. This allowed us to assess how sensitive the model was to the capacity of the height embedding and whether performance gains depended on a specific architectural choice.

\subsubsection{Processing time}
In the previous analysis, the processing time covered preprocessing and network inference on point clouds that had already been created and filtered. Although this gives an indication of the neural network throughput, it does not reflect the complete throughput of PointRAFT on an operational harvesting system, because that also requires point cloud creation and point cloud filtering (the steps explained in Section~\ref{point_cloud_creation}). To include this more practical perspective and to broaden the interpretation towards different computing devices, we conducted a more detailed processing time analysis of the complete pipeline on five computing devices. These devices ranged from high-end GPU-equipped PCs and laptops to embedded devices that one would typically find in a real implementation on a harvester. The first computing device was the laptop described in Section~\ref{imaging_system}. The analysis was further conducted on a GALLERIA ZA9R desktop PC and an Alienware m15 R7 laptop, as well as on two edge devices: a ZOTAC ZBOX mini PC and a reComputer J4012 embedded computer.

On each device, all 5,254 test samples were processed after 20 warm-up runs. For each test sample, the analysis measured the time required for loading the RGB and depth images with the precomputed instance masks, point cloud creation, DBSCAN filtering, PointRAFT preprocessing (the steps explained in Section~\ref{benchmark_analysis}), and PointRAFT network inference. Instance segmentation and the one-time model loading were excluded. The mean total time per sample was finally converted into a throughput, expressed as the number of samples processed per second.

\clearpage{}
\section{Results}
\label{results}

\subsection{Benchmark analysis}

Table~\ref{tab:benchmark_results} summarizes the benchmark results by potato cultivar, growing season, and weight class, as well as for the entire test set, calculated per point cloud and per tuber. Per point cloud, PointRAFT achieved the lowest MAE of 11.7~g and the lowest RMSE of 17.1~g, followed by DGCNN with an MAE of 13.4~g and an RMSE of 18.4~g. The remaining networks achieved MAEs between 14.3~g (PTv3-C) and 16.5~g (PointNet). Relative to PointNet++, which shares its backbone with PointRAFT but lacks the object height embedding, PointRAFT reduced MAE by 4.1~g (25.9\%) and RMSE by 4.5~g (20.8\%). Per tuber, PointRAFT again achieved the lowest MAE of 10.2~g and the lowest RMSE of 15.2~g, followed by DGCNN with an MAE of 11.6~g and an RMSE of 15.8~g. Relative to PointNet++, PointRAFT reduced the per-tuber MAE by 4.4~g (30.1\%) and RMSE by 4.4~g (22.4\%). The Wilcoxon signed-rank test showed that the errors of PointRAFT were significantly lower than those of all six other networks ($p < 0.05$), with the smallest difference for DGCNN ($p = 0.005$). This confirms that the improvements of PointRAFT also hold for the weight estimates of individual tubers.

\begin{table}[hbt!]
    \captionsetup{justification=raggedright,singlelinecheck=false}
    \caption{Weight regression results of the seven point cloud regression networks per potato cultivar, growing season, and weight class, and for the entire test set, calculated per point cloud and per tuber. Results are given as the mean absolute error, with the root mean square error in parentheses. The lowest mean absolute error and the shortest processing time are printed in bold. PTv3-C denotes our compact adaptation of Point Transformer V3.}
    \centering
    \scriptsize
    \setlength{\tabcolsep}{5pt}
    \begin{tabular}{l l c c@{\hspace{15pt}} c c c c c c c}
        \hline
        & 
        & Potato & Point & \multicolumn{7}{c}{Mean absolute error \rmse{root mean square error} [g] $\downarrow$} \\\cline{5-11}
        \multicolumn{2}{l}{Subset comparison} & tubers & clouds & PointNet & PointNet++ & DGCNN & PointMLP & RandLA-Net & PTv3-C & PointRAFT\\
        \hline
        \multirow{4}{1.0cm}{\raggedright Potato \\ cultivar} & Sayaka & 59 & 1,670 & 13.9 \rmse{19.3} & 14.0 \rmse{18.9} & 11.2 \rmse{15.1} & 11.5 \rmse{16.1} & 13.2 \rmse{18.4} & 13.1 \rmse{17.8} & \textbf{10.3} \rmse{14.8}\\
        & Kitahime & 98 & 3,144 & 16.7 \rmse{23.1} & 15.5 \rmse{21.6} & 14.1 \rmse{19.4} & 16.0 \rmse{23.3} & 16.2 \rmse{21.9} & 14.5 \rmse{20.8} & \textbf{11.6} \rmse{16.7}\\
        & Corolle & 8 & 222 & 23.7 \rmse{27.4} & 24.0 \rmse{30.1} & 14.8 \rmse{19.6} & 14.6 \rmse{18.7} & 18.8 \rmse{21.9} & \textbf{13.4} \rmse{17.6} & 15.5 \rmse{21.9}\\
        & Haruka & 7 & 218 & 26.3 \rmse{31.0} & 24.1 \rmse{28.9} & 19.3 \rmse{23.7} & 20.5 \rmse{25.9} & 25.2 \rmse{29.8} & 21.1 \rmse{24.4} & \textbf{19.2} \rmse{29.7}\\ \hline
        
        \multirow{3}{1.0cm}{\raggedright Growing \\ season} & 2023 & 80 & 2,271 & 16.3 \rmse{21.8} & 16.2 \rmse{21.5} & 12.5 \rmse{16.8} & 13.1 \rmse{18.1} & 15.2 \rmse{20.4} & 14.0 \rmse{18.6} & \textbf{11.4} \rmse{17.3}\\
        & 2024 & 49 & 1,485 & 18.4 \rmse{26.2} & 17.1 \rmse{24.5} & 14.9 \rmse{21.7} & 17.3 \rmse{25.8} & 16.7 \rmse{23.4} & 15.9 \rmse{23.2} & \textbf{12.4} \rmse{17.7}\\
        & 2025 & 43 & 1,498 & 15.1 \rmse{19.8} & 13.8 \rmse{18.4} & 13.3 \rmse{17.1} & 14.4 \rmse{20.5} & 15.7 \rmse{20.3} & 13.1 \rmse{18.2} & \textbf{11.2} \rmse{16.2}\\ \hline
        
        \multirow{6}{1.0cm}{\raggedright Weight \\ class [g]} & 0--100 & 56 & 1,834 & 11.2 \rmse{14.0} & 9.2 \rmse{11.9} & 7.9 \rmse{9.9} & 8.3 \rmse{10.9} & 9.7 \rmse{12.0} & 7.5 \rmse{9.8} & \textbf{5.9} \rmse{7.8}\\
        & 100--200 & 53 & 1,613 & 14.4 \rmse{19.5} & 15.2 \rmse{19.7} & 13.5 \rmse{17.3} & 13.2 \rmse{18.7} & 14.9 \rmse{19.3} & 13.9 \rmse{17.7} & \textbf{10.4} \rmse{14.5}\\
        & 200--300 & 40 & 1,131 & 20.4 \rmse{25.6} & 20.7 \rmse{26.2} & 16.5 \rmse{21.5} & 18.4 \rmse{23.6} & 20.1 \rmse{25.1} & 19.1 \rmse{24.2} & \textbf{15.2} \rmse{20.9}\\
        & 300--400 & 19 & 562 & 27.2 \rmse{34.9} & 25.1 \rmse{32.1} & \textbf{21.2} \rmse{27.1} & 27.4 \rmse{34.5} & 24.5 \rmse{31.4} & 24.6 \rmse{31.0} & 21.5 \rmse{25.7}\\
        & 400--500 & 2 & 53 & 56.6 \rmse{63.8} & 50.9 \rmse{56.3} & \textbf{40.6} \rmse{49.9} & 56.9 \rmse{63.1} & 53.5 \rmse{59.5} & 45.3 \rmse{55.7} & 49.6 \rmse{52.1}\\
        & 500--600 & 2 & 61 & 29.6 \rmse{33.2} & 22.1 \rmse{25.9} & 24.0 \rmse{27.2} & 23.9 \rmse{32.7} & 24.6 \rmse{29.6} & \textbf{19.1} \rmse{26.0} & 28.8 \rmse{37.1}\\ \hline\hline

        \multirow{2}{1.0cm}{\raggedright Test set} & Per cloud & \footnotesize 172 & \footnotesize 5,254 & \footnotesize 16.5 \rmse{22.6} & \footnotesize 15.8 \rmse{21.6} & \footnotesize 13.4 \rmse{18.4} & \footnotesize 14.7 \rmse{21.2} & \footnotesize 15.7 \rmse{21.3} & \footnotesize 14.3 \rmse{19.9} & \footnotesize \textbf{11.7} \rmse{17.1}\\
        & Per tuber & \footnotesize 172 & \footnotesize 5,254 & \footnotesize 15.8 \rmse{21.3} & \footnotesize 14.6 \rmse{19.6} & \footnotesize 11.6 \rmse{15.8} & \footnotesize 13.4 \rmse{19.4} & \footnotesize 14.4 \rmse{19.4} & \footnotesize 12.6 \rmse{17.6} & \footnotesize \textbf{10.2} \rmse{15.2}\\ \hline
        \multicolumn{4}{l}{\multirow{2}{*}{Average processing time [ms] $\downarrow$}} & \multirow{2}{*}{\footnotesize\textbf{4.5}} & \multirow{2}{*}{\footnotesize 6.1} & \multirow{2}{*}{\footnotesize 5.0} & \multirow{2}{*}{\footnotesize 17.6} & \multirow{2}{*}{\footnotesize 17.2} & \multirow{2}{*}{\footnotesize 8.2} & \multirow{2}{*}{\footnotesize 6.3}\\
        \multicolumn{4}{l}{} & & & & & & & \\ \hline\hline

    \end{tabular}
    \label{tab:benchmark_results}
\end{table}

For Sayaka and Kitahime, the two main cultivars in the test set, PointRAFT achieved the lowest MAE (10.3~g and 11.6~g) and RMSE (14.8~g and 16.7~g, respectively). Results were mixed for the two smaller cultivars: PointRAFT achieved the lowest MAE for Haruka (19.2~g), whereas PTv3-C outperformed PointRAFT for Corolle (13.4~g compared to 15.5~g), although both results were based on only seven and eight test tubers, respectively. Across the three growing seasons, PointRAFT achieved the lowest MAE in every season, ranging from 11.2~g to 12.4~g. The results for the 2025 growing season are particularly encouraging, because many tubers in that season were deformed by the hot and dry growing conditions. For the RMSE, PointRAFT also had the lowest values in 2024 and 2025, whereas in 2023 DGCNN achieved a slightly lower RMSE (16.8~g compared to 17.3~g).

For the different weight classes, PointRAFT achieved the lowest MAE in the three lightest classes up to 300 grams. For the 300--400~g class, DGCNN and PointRAFT performed similarly (MAE of 21.2~g and 21.5~g, respectively), with PointRAFT achieving the lowest RMSE of 25.7~g. In the two heaviest weight classes, DGCNN (400--500~g) and PTv3-C (500--600~g) achieved the lowest MAEs, while PointRAFT ranked lower. Each of these two classes, however, contained only two test tubers. A trend observed at all regression networks is that the absolute errors increased with tuber weight. This trend may have two explanations. First, heavier tubers were underrepresented in the training data, providing fewer examples to learn from. Second, heavier tubers typically had less regular shapes, which made it harder to infer the parts that could not be observed from the top-view camera perspective.

Regarding the average processing time, PointRAFT required 6.3~ms per point cloud and ranked fourth after PointNet (4.5~ms), DGCNN (5.0~ms), and PointNet++ (6.1~ms). The difference with the fastest network, PointNet, was only 1.8~ms, and compared to PointNet++, PointRAFT required just 0.2~ms more, indicating that the object height embedding added a negligible computational cost. Notably, DGCNN was the second-fastest network and also achieved the second-best performance, making it a potentially interesting alternative to PointRAFT.

\begin{figure}[htbp]
  \centering
    \includegraphics[width=1\textwidth]{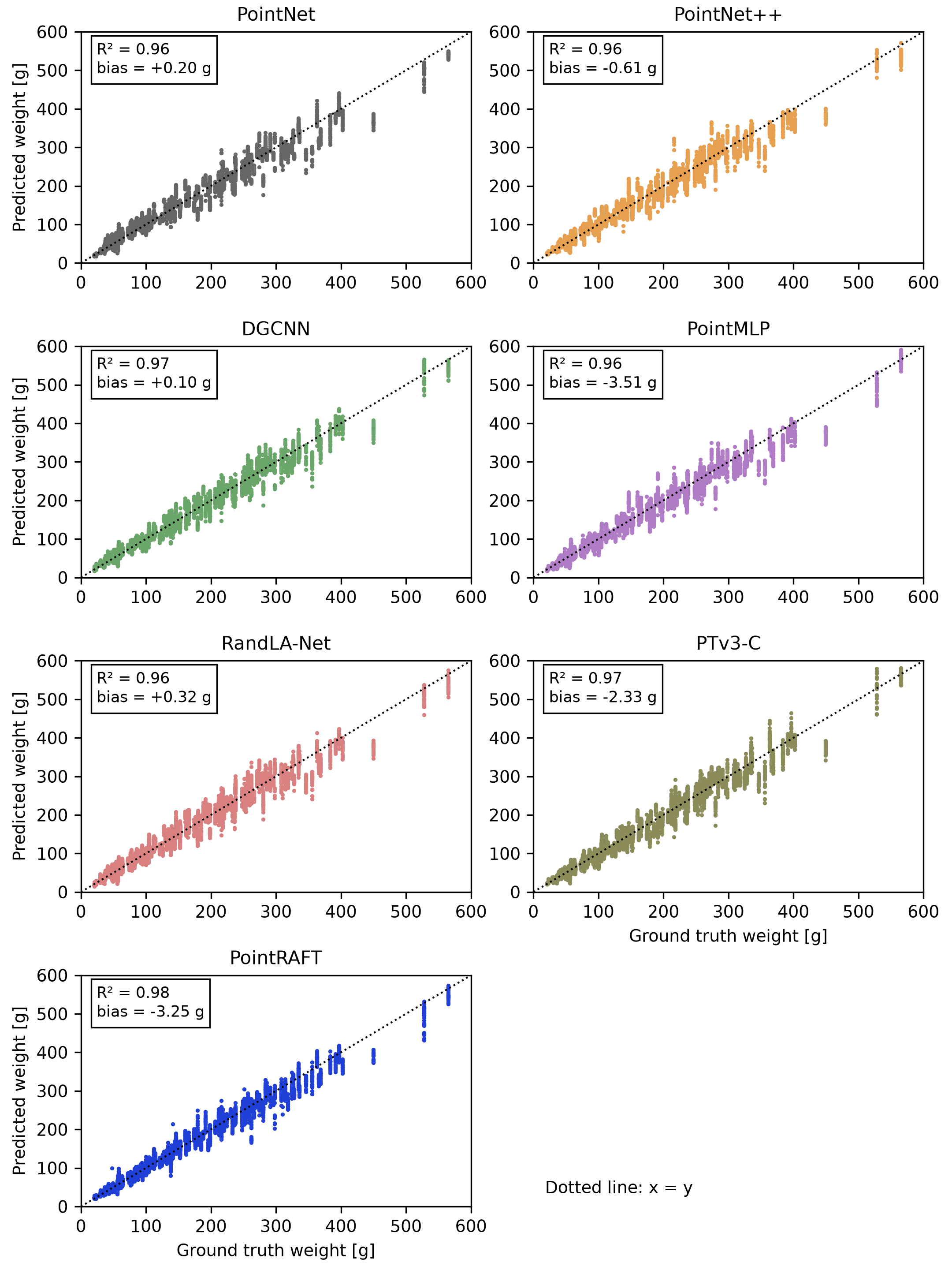}
    \caption{Predicted versus ground truth weights of the seven point cloud regression networks on the 5,254 test point clouds. Each panel reports the coefficient of determination (R²) and the bias, defined as the mean signed error. The dotted line indicates perfect agreement (x = y).}
    \label{fig:scatter_plot}
\end{figure}

Figure~\ref{fig:scatter_plot} shows the predicted weights against the ground truth weights for the entire test set. PointRAFT achieved the highest $R^2$ of 0.98, followed by DGCNN and PTv3-C (both 0.97) and the remaining four networks (all 0.96). Despite its highest $R^2$, PointRAFT showed a negative bias of $-3.25$~g, which was mainly caused by a systematic underestimation of tubers heavier than 200~g (bias of $-9.3$~g), as also shown by their predictions lying mostly below the dotted line. For tubers up to 200~g, the bias was negligible ($-0.06$~g). DGCNN had the smallest bias of all networks ($+0.10$~g), indicating that its predictions showed no systematic over- or underestimation. Another trend visible in Figure~\ref{fig:scatter_plot} is that the predicted weight varied across the different viewpoints of the same tuber, resulting in vertical bands of predictions for each tuber as it moved along the conveyor belt. 

\begin{figure}[hbt!]
    \centering
    \includegraphics[width=\textwidth]{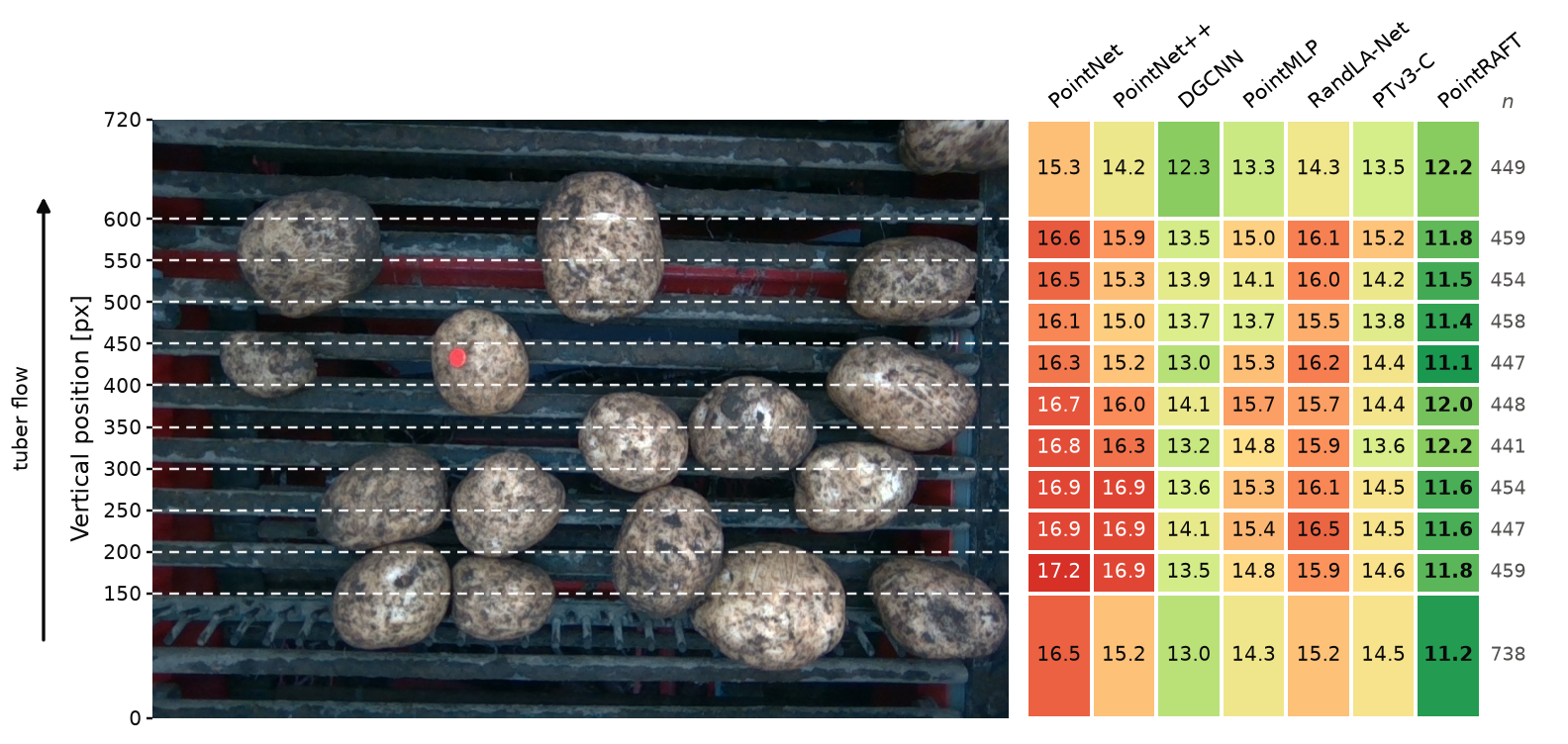}
    \caption{Mean absolute error (MAE) of the seven point cloud regression networks per horizontal image region, calculated over the entire test set (5,254 point clouds). The number of point clouds per region is given by $n$. The cell colors range from green (lowest MAE, 11.1~g) to red (highest MAE, 17.2~g) across all regions and networks, and the lowest MAE per region is printed in bold. The background image indicates the region boundaries; the tracked tuber in this example is the one with a red thumbtack in the 400--450 pixel region.}
    \label{fig:performance_regions}
\end{figure}

This viewpoint dependency is examined further in Figure~\ref{fig:performance_regions}, which shows the MAE of each network per horizontal image region. PointRAFT achieved the lowest MAE in all eleven regions, with the lowest value of 11.1~g in the 400--450 pixel region. Across the image, the MAE of PointRAFT varied by only 1.1~g, ranging from 11.1~g to 12.2~g. Although this variation was small, performing the regression in a single region with the lowest MAE would reduce the computational load, making the 400--450 pixel region the most suitable region for PointRAFT.

\subsection{Ablation study}

The ablation study (Table~\ref{tab:ablation_study}) shows that every tested ablation increased the MAE of PointRAFT, with data augmentation having the largest effect: disabling it increased MAE by 16.8\%. By randomly rotating, flipping, shearing and jittering the training point clouds, data augmentation exposed the network to a wider variety of tuber shapes and orientations than present in the original data. This helped PointRAFT learn weight-relevant shape features rather than memorizing the specific training samples, which improved its generalization performance on unseen tubers.

Input point density determined the trade-off between prediction accuracy and total inference time. Reducing the input to 512 points made PointRAFT 14.8\% faster (5.4 ms instead of 6.3 ms per point cloud), but this speed gain came at the cost of performance, with MAE increasing by 5.8\% and RMSE by 3.1\%. Doubling the input to 2048 points had the opposite effect on speed, increasing the total inference time by 44.6\% to 9.1~ms, without any gain in performance: MAE and RMSE increased by 4.0\% and 5.2\%, respectively. Training with 1024 points therefore achieved the best balance between prediction accuracy and total inference time, with an average of 6.3 ms per point cloud.

The choice of loss function had a clear effect on the accuracy of the weight estimates, with both alternative losses performing worse than the default Smooth L1 loss. Replacing the Smooth L1 loss with the less outlier-sensitive L1 loss increased MAE by 5.2\% and RMSE by 3.7\%, while the more outlier-sensitive MSE loss increased MAE by 8.5\% and RMSE by 4.6\%. The Smooth L1 loss thus offered the best compromise, combining the robustness of the L1 loss to large errors with the smooth gradients of the MSE loss for small errors.

The width of the height embedding MLP had a consistent effect on the accuracy of the weight estimates, with every alternative width performing worse than the default [1, 8, 8] configuration. The clearest trend appeared in MAE, which increased most for the two extreme configurations: the narrowest embedding of [1, 2, 2] increased MAE by 10.1\% and the widest embedding of [1, 64, 64] by 10.4\%. The intermediate widths of [1, 4, 4], [1, 16, 16], [1, 32, 32] and [1, 48, 48] resulted in smaller MAE increases of 5.8\% to 9.3\%. The RMSE followed this pattern less strictly, increasing by 3.9\% to 7.0\% for all configurations except [1, 4, 4], for which it remained essentially unchanged ($-$0.6\%). Intuitively, a very narrow embedding may be too limited to represent the relation between tuber height and weight, whereas a very wide embedding adds many parameters to encode a single scalar value, which may make it harder to learn and to balance against the geometric features. A compact height embedding of moderate width therefore appears sufficient to encode tuber height, with [1, 8, 8] providing the best results in this study.

\begin{table}[htbp]
\caption{Performance metrics and total inference time for the eleven studied ablations relative to the baseline performance of PointRAFT. The largest increase in MAE, RMSE and total inference time is printed in bold.}
\centering
\scriptsize
\begin{tabular}{llcccccccc}
    \hline
    & & \multicolumn{2}{c}{MAE [g] $\downarrow$} & & \multicolumn{2}{c}{RMSE [g] $\downarrow$} & & \multicolumn{2}{c}{Time [ms] $\downarrow$}\\
    \cline{3-4} \cline{6-7} \cline{9-10}
    Ablation & Category & abs & rel & & abs & rel & & abs & rel\\ 
    \hline
    PointRAFT & Baseline & 11.7 & - & & 17.1 & - & & 6.3 & -\\ \hline \hline
    1024 $\rightarrow$ 512 & Input points & 12.3 & +5.8\% & & 17.7 & +3.1\% & & 5.4 & $-$14.8\%\\
    1024 $\rightarrow$ 2048 & Input points & 12.1 & +4.0\% & & 18.0 & +5.2\% & & \textbf{9.1} & \textbf{+44.6\%}\\
    No data augmentation & Network training & \textbf{13.6} & \textbf{+16.8\%} & & \textbf{19.3} & \textbf{+12.6\%} & & 6.2 & $-$0.9\%\\
    Smooth L1 $\rightarrow$ L1 & Loss function & 12.3 & +5.2\% & & 17.8 & +3.7\% & & 6.2 & $-$1.7\%\\
    Smooth L1 $\rightarrow$ MSE & Loss function & 12.7 & +8.5\% & & 17.9 & +4.6\% & & 6.2 & $-$2.1\%\\
    Height embed. [1,2,2] & Network architecture & 12.8 & +10.1\% & & 17.8 & +3.9\% & & 6.2 & $-$0.9\%\\
    Height embed. [1,4,4] & Network architecture & 12.3 & +5.8\% & & 17.0 & $-$0.6\% & & 6.2 & $-$1.8\%\\
    Height embed. [1,16,16] & Network architecture & 12.7 & +9.3\% & & 18.2 & +6.1\% & & 6.1 & $-$3.3\%\\
    Height embed. [1,32,32] & Network architecture & 12.5 & +6.8\% & & 18.3 & +7.0\% & & 6.3 & $-$0.2\%\\
    Height embed. [1,48,48] & Network architecture & 12.4 & +6.7\% & & 18.1 & +5.4\% & & 6.2 & $-$1.9\%\\
    Height embed. [1,64,64] & Network architecture & 12.9 & +10.4\% & & 18.2 & +6.3\% & & 6.2 & $-$1.5\%\\
    \hline \hline
\end{tabular}
\label{tab:ablation_study}
\end{table}

\subsection{Processing time}
The processing times in Tables~\ref{tab:benchmark_results} and \ref{tab:ablation_study} only covered point cloud preprocessing and network inference. Table~\ref{tab:inference_times} additionally includes image loading, point cloud creation, and DBSCAN filtering, and therefore represents the complete processing pipeline.

On all devices, image loading and DBSCAN filtering were by far the most time-consuming steps, accounting for 89 to 98\% of the total processing time. With these steps included, the throughput was 6.0 samples per second on the Lenovo Legion Pro, 5.6 on the GALLERIA ZA9R, 2.0 on both the Alienware m15 R7 and the ZOTAC ZBOX, and 1.5 on the reComputer J4012. An important nuance is that the cameras on a harvester stream their RGB and depth frames directly, so that loading the images from disk is not required in a practical implementation. Without image loading, the throughput would increase to approximately 7.1 samples per second on the Lenovo Legion Pro, 6.9 on the GALLERIA ZA9R, 2.2 on both the Alienware m15 R7 and the ZOTAC ZBOX, and 1.6 on the reComputer J4012. Among the two edge devices, the ZOTAC ZBOX was the most promising, as PointRAFT preprocessing and network inference together took only 10.5~ms on this device. However, its throughput was largely limited by DBSCAN filtering (445.2~ms), so a faster filtering method would be essential to make it suitable for high-throughput operation on a potato harvester.

\begin{table}[htbp]
\centering
\caption{Hardware configurations and mean stage-wise processing time and throughput.}
\label{tab:inference_times}
\scriptsize
\resizebox{\textwidth}{!}{%
\begin{tabular}{lccccc}
\hline
& \multicolumn{2}{c}{Laptop} & \multicolumn{1}{c}{PC} & \multicolumn{2}{c}{Edge device} \\
\cmidrule(r){2-3}\cmidrule(lr){4-4}\cmidrule(l){5-6}
Device & Lenovo Legion Pro & Alienware m15 R7 & GALLERIA ZA9R & ZOTAC ZBOX & reComputer J4012 \\
\hline
OS & Ubuntu 24.04 & Windows 11 & Manjaro Linux 6.18 & Debian 13 & JetPack 5.1.3 \\
CPU & Intel i9-13900HX & AMD Ryzen 7 6800H & AMD Ryzen 9 5950X & Intel i7-7700HQ & 8-core ARMv8 \\
RAM & 32 GB & 16 GB & 128 GB & 16 GB & 16 GB \\
GPU & RTX 4090 (M) & RTX 3070 Ti (M) & RTX 3090 & GTX 1080 & Jetson Orin \\
VRAM & 16 GB & 8 GB & 24 GB & 8 GB & Shared 16 GB \\
CUDA & 12.8 & 12.8 & 12.8 & 12.6 & 11.4 \\
NVIDIA driver & 580.178.04 & 610.60 & 610.57.04 & 550.163.01 & L4T 35.5.0 \\
\hline
RGB and depth image loading [ms] & 25.4 & 50.0 & 31.7 & 50.4 & 62.0 \\
RGB-D image to point cloud [ms] & 0.5 & 2.0 & 0.6 & 1.0 & 3.5 \\
DBSCAN filtering [ms] & 135.1 & 389.3 & 128.9 & 445.2 & 571.7 \\
PointRAFT preprocessing [ms] & 2.4 & 24.8 & 3.2 & 3.9 & 10.0 \\
PointRAFT network inference [ms] & 3.5 & 29.8 & 12.8 & 6.6 & 36.4 \\
\hline
\textbf{Total measured time [ms]} & \textbf{166.9} & \textbf{496.0} & \textbf{177.2} & \textbf{507.0} & \textbf{683.7} \\
\textbf{Throughput (samples/s)} & \textbf{6.0} & \textbf{2.0} & \textbf{5.6} & \textbf{2.0} & \textbf{1.5} \\
\hline
\end{tabular}
}
\end{table}

\clearpage
\section{Discussion}
\label{discussion}

In this study, we hypothesized that PointRAFT's height embedding improves weight estimation from partial point clouds, compared to regression networks that rely on the point cloud geometry alone. Our results support this hypothesis. On the entire test set, PointRAFT achieved the lowest MAE and RMSE of all seven networks, both per point cloud and per tuber, and it achieved the lowest MAE in nine of the thirteen subsets. The most direct evidence came from the comparison with PointNet++, which shares its backbone with PointRAFT and differs only in the absence of the height embedding. Adding the height embedding reduced MAE by 25.9\% and RMSE by 20.8\%, while increasing the processing time by only 0.2~ms. The ablation study further showed that this improvement did not depend on a specific architectural choice: all six alternative embedding widths resulted in MAEs between 12.3~g and 12.9~g, which were still considerably lower than the 15.8~g of PointNet++. This indicates that the performance gain originated from the height information itself rather than from the exact configuration of the embedding. Finally, PointRAFT performed consistently across the three growing seasons, with MAEs between 11.2~g and 12.4~g, even though these seasons involved two different camera-to-belt distances and included the deformed tubers of 2025. This indicates that the height embedding remained effective for both camera setups and all growing conditions in our dataset, although generalization to new seasons or camera setups was not tested.

Despite these promising results, several limitations remain. First, only four potato cultivars were included, and two of them, Haruka and Corolle, had limited sample sizes. This restricts our ability to assess how well PointRAFT generalizes to cultivars with very different shapes. In particular, the height embedding may have been biased toward the dominant cultivars, Sayaka and Kitahime, which had similar length–width–height ratios. In practice, PointRAFT may therefore generalize less effectively to more elongated or more spherical tubers. Second, all growing seasons were included in the training set, and growing season, camera setup, and cultivar were partially linked. The generalization of PointRAFT to new growing seasons and camera setups therefore remains to be confirmed. Third, the dataset was imbalanced, with only 23 of the 172 test tubers weighing more than 300~g. All networks had their largest errors on these heavier tubers, and PointRAFT systematically underestimated tubers heavier than 200~g (bias of $-9.3$~g). Such a systematic underestimation is undesirable for yield monitoring, because each heavy tuber contributes considerably to the total yield. Fourth, most tubers in our dataset were not stacked, because the sampled tubers had to be tagged and remain visible as they moved through the camera's field of view. In real harvesting conditions, tuber stacking occurs more often. In such cases, using the conveyor belt as a height reference may lead to overestimation when stacked tubers are interpreted as single objects. Fifth, tuber orientation was not explicitly modeled, so tilting may have caused small height estimation errors. Finally, a direct comparison of PointRAFT with published potato weight estimation methods was difficult, because, to our knowledge, no other method regresses potato weight directly from partial point clouds.

Future research should address these limitations. First, a larger dataset with more cultivars and more heavy tubers could improve the generalization of PointRAFT. Data from additional growing seasons would also allow a leave-one-season-out evaluation, which provides a better test of generalization. Second, stacked tubers could be handled by incorporating visual context extracted from instance segmentation, allowing tuber height to be estimated relative to neighboring tubers rather than to the conveyor belt. Third, tilted tubers could be handled by providing the tuber orientation to the network as an additional embedding. Finally, a unified benchmark with a shared test set and evaluation protocol would enable direct comparisons between potato weight estimation methods. The public subset of our dataset \citep{wang2025} can serve as a starting point for such a benchmark.

Regarding the technical implementation of PointRAFT, there are also several potential improvements. First, the variation in predictions across multiple views of the same tuber suggests using a consistency loss to reduce differences between viewpoints. This could further lower MAE and RMSE. Second, the comparatively short PointRAFT regression stage allows multiple stochastic forward passes after point cloud preprocessing to estimate prediction uncertainty. Methods such as Monte Carlo dropout \citep{gal2016} are well suited for this and can be implemented with minimal changes. Third, DGCNN may be a promising alternative backbone for the height embedding, as it achieved the second-best accuracy and the second-shortest processing time in our benchmark. Unlike PointNet++, which aggregates features within fixed spatial radii, DGCNN dynamically recomputes its neighborhood graph in the learned feature space, allowing it to relate points that are spatially distant but geometrically similar. This property may be beneficial for partial point clouds, in which the visible surface has to be related to the overall tuber shape. Moreover, PointRAFT's height embedding added only 0.2~ms to the processing time of PointNet++, so that a DGCNN with the same embedding would likely remain faster than PointRAFT. Such a network could therefore combine the benefits of both approaches. To facilitate this extension, our code, network weights, and a subset of our dataset are publicly available.

The practical implementation of PointRAFT on a harvester depends on the throughput of the complete processing pipeline. Although PointRAFT's preprocessing and network inference required only 6.3~ms per point cloud, the complete pipeline was mainly governed by image loading and DBSCAN filtering. Together, these two steps accounted for 89 to 98\% of the total processing time on all devices. On an operational harvester, image loading can be omitted, because the camera streams its RGB and depth frames directly. RGB-D cameras that generate point clouds onboard could further simplify the pipeline, although the time gain would be small, as point cloud creation required at most 3.5~ms. Even without image loading, the throughput would remain limited to 1.6 to 7.1 samples per second, because DBSCAN filtering alone accounted for 73 to 88\% of the total processing time.

We therefore consider DBSCAN filtering unsuitable for practical implementation and recommend faster filtering methods. One option is to filter the masked depth image instead of the point cloud. Because the depth image is already arranged in pixels, a connected component analysis on depth discontinuities can remove small and spatially separated fragments. This is expected to require only a fraction of the time of DBSCAN. Other options are statistical outlier removal \citep{rusu2008}, which has already been applied in RGB-D-based weight estimation of bananas \citep{fu2025}, or the filtering method described in \citet{blok2021}. These faster methods may, however, filter less accurately than DBSCAN, which could affect the weight regression. It is thus important to investigate how sensitive PointRAFT is to the choice of filtering method.

When PointRAFT is implemented as part of a yield monitoring system, it is also important to consider the impact of upstream and downstream tasks. In our study, the point clouds were extracted using manually annotated masks. In an operational system, these masks have to be predicted by an instance segmentation algorithm. In practice, these predictions may contain errors, which can lead to noisy point clouds and reduced prediction accuracy, indicating that the sensitivity of PointRAFT to imperfect masks should be investigated. Regarding the downstream tasks, the weight of individual tubers is not the output that a farmer needs to interpret the yield. The individual weight estimates therefore have to be combined into a spatial yield map, which requires integration with a global navigation satellite system (GNSS) to geo-reference predictions and visualize yield variability across the field.

Although our study focused on potato tubers, PointRAFT is not limited to this application. Its modular design, user-friendly code base, and comparatively short regression stage make it suitable for other agricultural tasks, such as size estimation of occluded fruits and vegetables. For these tasks, however, the height embedding might need to be revisited. On a potato harvester, tuber height can be derived from the fixed distance between the camera and the conveyor belt, but such a fixed reference distance is absent in most other settings. In tree canopies and greenhouse crops, for example, fruits are scattered at different depths, so that their extent along the viewing direction of the camera cannot be derived from a known background. In such cases, the height embedding could instead encode a range of fruit heights, measured beforehand on a representative sample of harvestable fruits. This range would provide prior geometric knowledge of the crop, although less specific than the height of each individual tuber. Whether such a coarser prior also improves regression performance remains to be investigated.

\clearpage
\section{Conclusions}
\label{conclusions}
In this study, we introduced PointRAFT, a high-throughput point cloud regression network capable of directly predicting tuber weight from partial point clouds. Its key architectural novelty was the object height embedding, which exploits the approximate tuber height as an additional geometric cue. On a test set of 5,254 point clouds, PointRAFT achieved an MAE of 11.7~g and an RMSE of 17.1~g, outperforming six point cloud regression networks without a height embedding. Compared to PointNet++, which shares its backbone with PointRAFT, the height embedding reduced the MAE by 25.9\%, confirming our hypothesis that this geometric cue improves weight estimation from partial point clouds. With a combined preprocessing and inference time of 6.3~ms per point cloud, PointRAFT enables real-time weight estimation of fast-moving tubers on a potato harvester, provided that the surrounding pipeline uses a faster point cloud filtering method. PointRAFT's accurate predictions provide farmers with actionable insights to optimize planting, fertilization, and harvesting in future seasons, helping to increase yield, reduce losses, and better meet market demand.

{\clearpage}

\section*{CRediT authorship contribution statement}
Pieter M. Blok: conceptualization, methodology, software, data curation, writing - original draft; Haozhou Wang: software, data curation, resources, writing - review \& editing; Hyun Kwon Suh: supervision, writing - review \& editing; Peicheng Wang: data curation, resources; James Burridge: conceptualization, methodology, data curation, writing - review \& editing; Wei Guo: conceptualization, methodology, funding acquisition, project administration, supervision, writing - review \& editing.

\section*{Funding}
This study is funded by the Sarabetsu Village ``Endowed Chair for Field Phenomics'' project in Hokkaido, Japan.

\section*{Acknowledgements}
We would like to thank Okada Farm for providing the potato fields on which we sampled the potatoes and acquired the RGB-D images. We thank Minato Miyauchi, Kyosuke Miyamoto, Kunihiro Tanaka, Xinbin Zhuang, Shuai Xiang, Ting Jiang, Sylvain Grison, Yuto Imachi and Irena Drofová for their help with the image acquisition and ground truth measurements. 

\section*{Declaration of generative AI and AI-assisted technologies in the manuscript preparation process}
During the preparation of this work, the author(s) used ChatGPT to support software debugging and to assist with language refinement. All content was subsequently reviewed and edited by the author(s), who take(s) full responsibility for the final published article.

\section*{Data availability}
A subset of the dataset used in this study is publicly available through the 3DPotatoTwin repository on Hugging Face at \url{https://huggingface.co/datasets/UTokyo-FieldPhenomics-Lab/3DPotatoTwin/tree/main}. The source code, pretrained PointRAFT model weights, and instructions for training and testing the model are publicly available at \url{https://github.com/pieterblok/pointraft}.

\section*{Declaration of competing interest}
The authors declare that they have no known competing financial interests or personal relationships that could have appeared to influence the work reported in this paper.

{\clearpage}

\bibliographystyle{apalike}
\bibliography{references}

\end{document}